\documentclass[11pt]{article}
\usepackage{coling2018}
\usepackage{times}
\usepackage{url}
\usepackage{latexsym}

\usepackage{algorithm}
\usepackage{algpseudocode}
\usepackage{multirow}
\usepackage{footnote}
\usepackage{xcolor}
\usepackage{colortbl}
\usepackage{amsmath,lipsum}
\usepackage{amsfonts}
\usepackage{graphicx}
\usepackage{tablefootnote}

\usepackage{wrapfig}

\usepackage[absolute]{textpos}

\usepackage{subcaption}

\usepackage{tikz}
\newcommand*\circled[1]{\tikz[baseline=(char.base)]{
            \node[shape=circle,draw,inner sep=0.5pt] (char) {#1};}}

\usepackage{enumitem}
\usepackage[skip=0pt]{caption}

\title{Location Name Extraction from Targeted Text Streams \\ using Gazetteer-based Statistical Language Models}

\author{Hussein S. Al-Olimat, Krishnaprasad Thirunarayan, Valerie L. Shalin and Amit P. Sheth \\
Kno.e.sis Center, Wright State University, Dayton, OH \\
\texttt{\{hussein;tkprasad;valerie;amit\}@knoesis.org} \\
}

\date{}

\begin{document}

\begin{textblock}{15}(0.5,0.3)
\small
\noindent \textcolor{red}{\textbf{Please cite:}} Hussein S. Al-Olimat, Krishnaprasad Thirunarayan, Valerie Shalin, and Amit Sheth. 2018. Location Name Extraction from Targeted Text Streams using Gazetteer-based Statistical Language Models. In Proceedings of the 27th International Conference on Computational Linguistics (COLING 2018), pages 1986-1997. Association for Computational Linguistics. Online: \url{https://www.aclweb.org/anthology/C18-1169.pdf}
\end{textblock}

\maketitle


\blfootnote{
    \hspace{-0.65cm}  
    This work is licensed under a Creative Commons 
    Attribution 4.0 International License.
    License details:
    \url{http://creativecommons.org/licenses/by/4.0/}
}


\begin{abstract}
Extracting location names from informal and unstructured social media data requires the identification of referent boundaries and partitioning compound names. Variability, particularly \emph{systematic} variability in location names \cite{carroll1983nameheads}, challenges the identification task. Some of this variability can be anticipated as operations within a statistical language model, in this case drawn from gazetteers such as OpenStreetMap (OSM), Geonames, and DBpedia. This permits evaluation of an observed $n$-gram in Twitter targeted text as a legitimate location name variant from the same location-context. Using $n$-gram statistics and location-related dictionaries, our Location Name Extraction tool (LNEx) handles abbreviations and automatically filters and augments the location names in gazetteers (handling name contractions and auxiliary contents) to help detect the boundaries of multi-word location names and thereby delimit them in texts.


We evaluated our approach on 4,500 event-specific tweets from three targeted streams to compare the performance of LNEx against that of ten state-of-the-art taggers that rely on standard semantic, syntactic and/or orthographic features. LNEx improved the average F-Score by 33-179\%, outperforming all taggers. Further, LNEx is capable of stream processing.\footnote{Data and the tool is available at \url{https://github.com/halolimat/LNEx}}
\end{abstract}


\section{Introduction}




In context-aware computing, location is a fundamental component that supports a wide-range of 
applications \cite{hazas2004location,licht2017location}. During natural disasters, location is crucial for situational awareness during disaster response \cite{son2008supporting}. When available, targeted streams of social media data are therefore particularly valuable for disaster response \cite{munro2011subword}\footnote{We define a \textit{targeted stream} as a set of tweets that has the potential to satisfy an event-related information need \cite{piskorski2013named} crawled using keywords and hashtags, to contextualize the event (e.g., ``\#ChennaiFloods'' for the floods in Chennai, India).}. For example, the tweet ``water level in Ganapathy Colony is around 2 m'' refers to a location. But, unless we know where ``Ganapathy Colony'' is, the water level data cannot enhance situational awareness and inform disaster response applications such as storm surge modeling/forecasting.



However, pragmatic influences on writing style shorten names to reduce redundant content in social media. We call this the \textit{location name contraction problem}. For example, ``Balalok School'', appears in the Chennai flood tweets in contrast to the full gazetteer name, ``Balalok Matriculation Higher Secondary School''.
\newcite{carroll1983nameheads} examined the complex phenomenon of alternate name forms (called Nameheads). He distinguishes between four shortening processes: (1) Appellation Formation, (2) Explicit Metonomy, (3) Category Ellipsis, and (4) Location Ellipsis. 


Appellation Formation occurs when, for example, the author refers to the location name ``The Erie Canal'' as ``The Canal''. 
People may also refer to the only airport in the affected area as just ``The Airport''. Referring to ``University of Michigan'' as ``Michigan'' is an example of Explicit Metonomy. Common ground or shared understanding between the author and the recipient establishes the referent \cite{resnick1991perspectives}. 
Both appelation formation and metonomy pose \emph{disambiguation} problems, and require context such as the author's location to resolve. 
%
%
%
%

In contrast, Category Ellipsis and Location Ellipsis pose \emph{delimitation} problems that resolve with a statistical language model. Category ellipsis occurs when the author strips words related to the location category (e.g., any of the intermediate tokens inside ``Balalok Matriculation Higher Secondary School'' or ``City'' from ``Houston City'' to become ``Houston''). Location Ellipsis occurs when an author drops the specific location reference in the location name (e.g., when ``New York Yankee Stadium'' becomes ``Yankee Stadium'' or ``Cars India - Adyar'' becomes ``Cars India''). 



This distinction between delimitation and disambigution is important in the location extraction literature. Entity delimitation
is typically the first step of location extraction, to identify the boundaries of a location mention in the text. To address this problem, previous research \cite{liu2014automatic,malmasi2015location,hoang2018location} has applied both syntactic heuristics (using lexical cues, e.g., ``\textit{in} New Orleans'') as well as semantic heuristics (i.e., content-based, for different types of locations such as \emph{buildings} and \emph{streets}). These heuristics 
have serious limitations, such as failing to delimit metonyms and location names that begin sentences (i.e., outside locative expressions, e.g., ``\textit{New Orleans} is flooded'') and they cannot assist in hashtag segmentation (needed to extracted locations from hashtags). Moreover, simply identifying a location name still leaves open the problem of linking the entity to a corresponding gazetteer record for geocoding. Simple fixed phrase matching with gazetteers entries, as in \cite{middleton2014real,malmasi2015location}, solves the linking problem, but remains vulnerable in two respects. 
With simple fixed phrase matching, the tendency for authors to shorten names while the gazetteers extend names, creates conflicting conditions causing poor recall. On the other hand, simply relaxing matching criteria exacerbates the disambiguation problem.

To address delimitation, we treat location names as a sequence of ordered words known as \textit{collocations} \cite{manning1999foundations}. Collocations are neither strictly compositional nor always atomic. We cannot identify them with grammatical rules, and fixed phrase matching is not reliable for longer names. Fortunately, the gazetteer provides a resource to establish region-specific naming regularities. Given a region-specific gazetteer, which retains the same location-context as the text, we can construct a statistical model of the \textit{token sequences} it contains. However, current gazetteers are overly specific in two respects. First, consistent with \cite{carroll1983nameheads}, they do not represent category ellipsis and location ellipsis. To mimic these processes, we judiciously apply a skip-gram method to token sequences in the gazetteers, thereby including, for example, ``Balalok School'' as a variant of the complete name. Second, we eliminate auxiliary or ambiguous gazetteers content (e.g., ``George, Washington'') that would otherwise threaten recall.

Here we answer the research question: \textit{\textbf{Can we accurately and rapidly spot location mentions in text solely relying on a statistical language model synthesized from augmented and filtered region-specific gazetteers?}} Although LNEx works on a targeted Twitter stream collected using event-specific keywords, it does not rely on rarely available tweets geo-coordinates and it does not need any supervision (i.e., training data). It is well-suited for stream processing, and needs only freely available data. Our contributions include:

\begin{enumerate}[topsep=0pt,noitemsep]
  \item A method for preparing high-quality gazetteers from online open data, such as OSM, Geonames, and DBpedia, and deriving a language model from them; and a comprehensive analysis of the contribution of gazetteer quality to overall performance. 
  
  \item A referent corpus representing the full scope of location name extraction challenges and a challenge-based categorization of place names found in the corpora resulting from targeted streams. We annotate three different Twitter streams from flooding events in three different locations: the 2015 Chennai flood, the 2016 Louisiana flood, and the 2016 Houston flood, for our own evaluation and also for use by others. 
  
  \item A demonstration that LNEx convincingly outperforms commercial-grade NER and Twitter-specific tools with at least a 33\% improvement on average F-Score. Examples reveal the  true challenges of location name extraction and the locus of tool failure in the face of these challenges.
\end{enumerate}

LNEx provides the foundation for localizing information, and with increased availability of open data, we expect our approach based on region-specific knowledge to be widely applicable in practice.


\section{The LNEx Method}


We discuss the details of LNEx in four subsections. First, we present the general idea of statistical inference via n-gram models;
the core of LNEx is a statistical language model consisting of a probability distribution over sequences of words (collocations) that represent location names in preexisting, region-specific gazetteers. 
Then we separately discuss several modifications to both gazetteers and text samples, including gazetteer augmentation and filtering, 
and tweet preprocessing.
Finally, we illustrate the full location analysis and matching process that reliably spots location names.
%
%






\subsection{Statistical Inference via $n$-gram Models}
\label{section:statsModel}

LNEx constructs an $n$-gram model from the collocations  that exist \textit{in the gazetteer} to determine the valid location names (LNs) that might appear in tweets. Given tweet content such as``texas ave is closed'', the model can then check the validity of one to $n$-grams. From the gazetteer, ``texas'' and ``ave'' are valid gazetteer unigrams but ``is'' and ``closed'' are not. Similarly, ``texas ave'' is a valid and preferred bigram (over two unigrams).

Specifically, as shown in Algorithm \ref{algo:lm}, we first tokenize all location names in the gazetteer to construct the $n$-gram model and then save the resulting lists of unigrams, bigrams, and trigrams (Lines 2-5). Next, for bigrams and trigrams, in Lines 6-9 we create conditional frequency distributions (CFD) to count the collocations (i.e., $c(\cdot)$ in equations \ref{equ:1}-\ref{equ:2}). Conditional probability distributions (CPD) are then constructed from the recorded $n$-grams using maximum likelihood estimation (MLE). We make the assumption that only the previous two words determine the probability of the next word (Markovian assumption of order two)\footnote{We found in our dataset that roughly 98\% of location mentions in tweets have less than three words.}. MLE 
assumes zero probability values to tokens missing from the gazetteers. \emph{This data sparsity problem is mitigated by augmenting the gazetteers with location name variants} (see Section \ref{sec:gazAug}).
In lines 11-13, we determine the validity of an $n$-gram (the string $s$) using the boolean function \texttt{\small VALID-N-GRAM} with the help of equations \ref{equ:1}-\ref{equ:n}, where $c(w_x^y) \equiv c(w_x w_{x+1} \dots w_y)$, $w_x^y$ is the collocation count (i.e., the occurrences of the consecutive words, $w_x$ to $w_y$), $P(w_z|w^{y}_{x})$ is the conditional probability of a word $w_z$ given previous collocation $w^{y}_{x}$, and the chain of probabilities $P_1$ (for unigrams), $P_2$ (for bigrams), and $P_3$ (for tri or larger grams).
%
%


\setlength{\abovedisplayskip}{-3pt}
\setlength{\belowdisplayskip}{5pt}

\setlength{\columnsep}{5pt}
\begin{wrapfigure}[0]{l}{0.45\textwidth}
\begin{align}
  P(w_z|w^{y}_{x})=\frac{c(w^{z}_{x})}{c(w^{y}_{x})} 
    \label{equ:1}\\
    P_1 = P(w_1^1) = \frac{c(w_1)}{\sum_{i=1}^{|\text{unigrams}|} c(w_i)}
    \label{equ:2}\\
    P_2 = P(w_1^2) = P_1 \times p(w_2 \mid w_1^1)
    \label{equ:3}\\
    P(w^{n}_1) = P_2 \times \prod_{i=3}^{n} P(w_i \mid w_{i-2}^{i-1}), n \geq 3
    \label{equ:n}
\end{align}

\end{wrapfigure}



\setlength{\columnsep}{15pt}
\begin{wrapfigure}[0]{r}{0.5\textwidth}
\vspace{-6.7cm}
    \begin{minipage}{0.5\textwidth}
\begin{algorithm}[H]
    \scriptsize
    \begin{algorithmic}[1]
        \Procedure{Compute-Model}{$Gazetteer$}

            \For{$ln \in Gazetteer$}
                \State $unigrams \leftarrow tokenize(ln)$;
                \State $bigrams, trigrams \leftarrow$ generate from $unigrams$;
            \EndFor

            \For{$n\text{-}grams \in [bigrams, trigrams]$}
                \State \textit{\textbf{CFD}} $\leftarrow$ create using $n\text{-}grams$;
                \State \textit{\textbf{CPD}} $\leftarrow$ create using \textit{\textbf{CFD}};
            \EndFor

        \EndProcedure
        \Statex
        \Procedure{Valid-N-Gram}{$string = s$}: boolean

          \State $w^{n}_1 = (w_1, \dots, w_n) \leftarrow tokenize(s)$;


          \Return $P(w^{n}_1) > 0$ \Comment{calculated using the equations (\ref{equ:1}-\ref{equ:n})};

        \EndProcedure
    \end{algorithmic}
    \caption{Language Model Generation}
    \label{algo:lm}
\end{algorithm}
\end{minipage}
  \end{wrapfigure}

\subsection{Gazetteer Augmentation and Filtering}
\label{sec:gazAug}

We faced two primary challenges when building our language model using raw gazetteers, which are not adequately explored in \cite{middleton2014real,weissenbacher2015knowledge}:

\begin{enumerate}[topsep=0pt,noitemsep]
  \item \textbf{Conditional Collocation Contractions}: Some atomic $n$-gram location names (collocations) cannot be shortened, e.g., ``New York''. However, contraction does preserve the meaning of some longer names, especially when the first and the last words denote a specific part and a generic part respectively, such as in ``Balalok School''.
  
  \item \textbf{Auxiliary and Spurious Content}: Gazetteer entries may contain extraneous content that can cause location matching to fail. Cleaning such entries improve matching reliability (see Table \ref{tbl:gazNoise}).
\end{enumerate}

To address these challenges, we exploited Category Ellipsis (for collocation contraction) and Location Ellipsis (for filtering the auxiliary content) as follows:

\begin{enumerate}[topsep=0pt,noitemsep]
\item \textbf{Skip-grams}: Given a location name $t_1 \dots t_n$, 
we retain $t_1$ and $t_n$ while varying $t_{2} \dots t_{n-1}$. To avoid adding ``City York'' as a legitimate variant of the location name ``City College of New York'', we require $t_n$ to be a location category name (e.g., building, road). Therefore, ``Balalok Matriculation Higher Secondary School'' 
generates \texttt{\small \{Balalok School, Balalok Secondary School, \dots\}}. This technique results in a small number of contractions that are either useful collocations or are too random to cause many false positives \cite{guthrie2006closer}.
\label{point:skipgram}



\item \textbf{Filtering}: To address bracketed auxiliary content, we compiled a generic list of phrases to remove specific words on a case-by-case basis (e.g., 1-2 in Table \ref{tbl:gazNoise}). The remaining bracketed names are deemed legitimate alternatives (e.g., 3-4 in Table \ref{tbl:gazNoise}). We treat hyphenated location names as Location Ellipsis and split them on the hyphen and add the two splits (e.g., 5 in Table \ref{tbl:gazNoise}). We expect that the majority of these location names represent a partonomy relationship where the hyphen may be read as a ``part of'' relation between split tokens. We do not add the second token as a variant when it already exists 
in the gazetteer on its own as location name entity (e.g., ``Hammond'' in ``Pilot - Hammond'').

\end{enumerate}



\setlength{\columnsep}{15pt}
\begin{wraptable}[6]{r}{0.41\textwidth}
\vspace{-0.5cm}
\def\arraystretch{1.5}
 \setlength{\tabcolsep}{0.42em}
 \centering
 \tiny
 \begin{tabular}{ l|l|l| }
    \cline{2-3}
    & \textbf{Content Description} & \textbf{Example Gazetteer Record}  \\
    \hline
    \multicolumn{1}{|c|}{1} & Descriptive Tags & (Private Road) \\
    \multicolumn{1}{|c|}{2} & Life-cycle/Status Tags & Little Rock School (historical) \\
    \multicolumn{1}{|c|}{3} & Alternative/Old Names & Scenic Road (Frontage Road) \\
    \multicolumn{1}{|c|}{4} & Acronyms & International House of Pancakes (IHOP) \\
    \multicolumn{1}{|c|}{5} & Hyphenations & Cars India - Adyar, Pilot - Hammond \\
    \hline
  \end{tabular}
  \caption{Extraneous text in raw gazetteers}
\label{tbl:gazNoise}
\end{wraptable}


These two methods augment and filter partial OSM, Geonames, and DBpedia gazetteers sliced from the original sources using a bounding box. Further, we can attach the metadata of the original location name to the generated variants. Moreover, by treating derived names as synonyms for existing names, we avoid creating additional demands on disambiguation or equivalencing. We add the derived, variant location name to the gazetteer as long as it does not collide with an existing location name. Additional filtering of proposed variants is required to prevent false alarms. Similar to the use case in \cite{weissenbacher2015knowledge,gelernter2013algorithm}, we compiled a list of 11,203 words including 678 inseparable bigrams, such as ``Building A'', as gazetteer stop words. This list also includes unusual location names (e.g., ``Boring'' in Maryland and ``Why'' in Arizona) and proper nouns (e.g., ``James'' in Mississippi) that could appear as non-location tokens. We then eliminate from all gazetteers the location names that overlap with our gazetteer stop words to reduce false positives.




\subsection{Tweet Preprocessing}
\label{sec:preprocesing}


To complement the gazetteer preprocessing, we also require potentially non-trivial tweet preprocessing.  We start by removing the retweet handles, URLs, non-ASCII characters, and all user mentions. Then, we tokenize tweets using TweetMotif's Twokenizer \cite{o2010tweetmotif}, which treats hashtags, mentions, and emoticons as a single token. We do not tokenize on periods (e.g., ``U.S.'').

\vspace{-0.2cm}

\paragraph{Hashtag Segmentation:}
In our datasets, on average, around 29\% of the hashtags include location names. Excluding hashtags used to crawl the data, around 17\% of the unique hashtags contain locations. As the number of locations in hashtags is significant, similar to \cite{malmasi2015location}, we adopted a statistical word segmentation algorithm to break hashtags 
for location spotting \cite{norvig2009natural}.




\vspace{-0.2cm}

\paragraph{Spelling Correction:}
We consider a tweet token as misspelled if it is an out-of-vocabulary token, where the vocabulary is gazetteer words and a large English vocabulary word list\footnote{\url{https://github.com/norrissoftware/words3}}. LNEx corrects all misspelled tokens using the Symmetric Delete Spelling Correction algorithm (SymSpell)\footnote{\url{https://github.com/wolfgarbe/symspell}} that is six orders of magnitude faster than Norvig's spelling corrector \cite{norvig2009natural}, which was used by \cite{gelernter2013cross} in their location extraction tool.
As we shall see, spelling correction has only a small influence on system accuracy.



\subsection{Extracting Location Names using LNEx}


%
%
%




\setlength{\columnsep}{15pt}
\begin{wrapfigure}[4]{r}{0.45\textwidth}
\vspace{-1.2cm}
    \centering
    \includegraphics[width=0.45\textwidth]{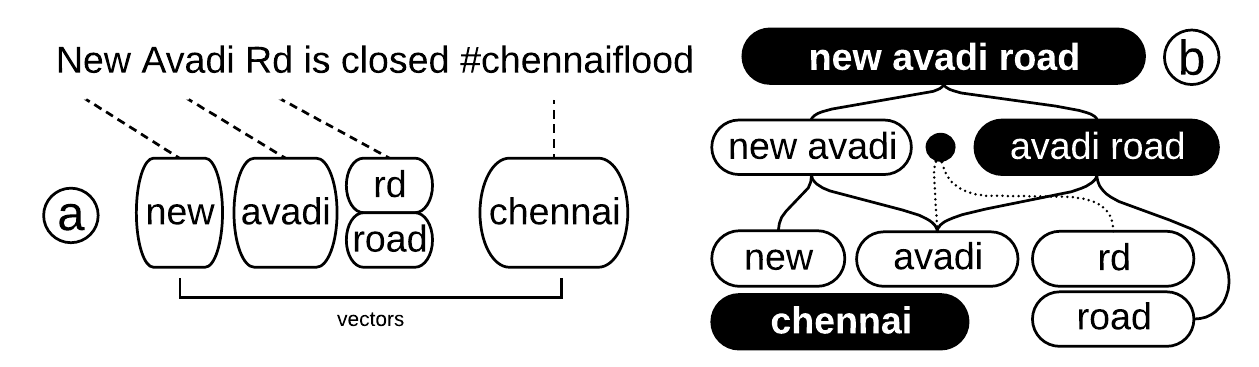}
    \caption{Extracting Locations using LNEx}
    \label{fig:LNExExample}
\end{wrapfigure}


After the modifications to gazetteers and texts, LNEx extracts locations as illustrated in Fig. \ref{fig:LNExExample}. In \circled{a}, LNEx reads the raw tweet text, preprocesses it (as in Section \ref{sec:preprocesing}) starting with case-folding. After tokenizing the tweet, the hashtag segmenter breaks hashtags into tokens. Later, stop words are used to split a tweet into consecutive word fragments where each tweet split of size $n$ can have zero to $n$ potential location names. We custom build the tweet stop list starting with around 890 words\footnote{\url{http://www.ranks.nl/stopwords}} excluding the gazetteer unigrams.
%
%
%
%
LNEx now takes each tweet split and converts each of its tokens into a vector of tokens $v$ using two dictionaries: the USPS street suffixes dictionary\footnote{\url{http://pe.usps.gov/text/pub28/28apc\_002.htm}} and the English OSM abbreviations dictionary\footnote{\url{wiki.openstreetmap.org/wiki/Name\_finder:Abbreviations}}. This adds possible expansions and abbreviations of a token (e.g., ``Rd'' to ``Road'', and vice versa). This overcomes the lexical variations between location mentions in tweets and their corresponding gazetteer entries.


In \circled{b}, the language model is used to find the valid $n$-grams from the Cartesian product of the consecutive vectors. It builds a bottom up tree for each tweet split starting from 1 to $n$-grams by gluing the consecutive tokens together if they represent a valid segment in the gazetteer. We improve the speed of the algorithm significantly by splitting the tweet and eliminating invalid $n$-grams. 
LNEx then selects a subset of valid $n$-grams from the tree; for the overlapping $n$-grams, we prefer the longest full mentions (e.g, ``New Avadi Road'' over ``Avadi Road'')
and keep both if they are of the same length. When full location names appear inside partial ones, we keep only full names 
(e.g., extracting ``Louisiana'' from ``The Louisiana''). 
%
%


\paragraph{Time and Space Complexities:}
LNEx extracts and links a full location mention to its corresponding gazetteer entry through a simple dictionary lookup that takes constant time $\mathcal{O}(1)$. The location extraction time is bounded by the time for creating the bottom up tree of tokens which takes $\mathcal{O}(|v|^s)$ where $|v|$ is the length of the longest vector of token synonyms (i.e., all the expansions and abbreviations of a token) and $s$ is the largest number of tokens with synonyms in a location name. 
Splitting the tweet into smaller fragments significantly lowers the asymptotic growth of the algorithm enabling stream processing. In practice, for our dictionaries and gazetteers, $|v| \leq 4$ and $s \leq 3$. So, a pessimistic upper bound on the number of candidates for each location (though rarely realized) is $4^3$. The space complexity of the method is bounded by the product of number of gazetteer entries, $L$, and the number of variants of a location name (Skip-gram method \ref{point:skipgram}), that is, $2^{m-2}$, where $m$ is the number of tokens in a location name. Effectively, the space complexity is $\mathcal{O}(L.2^{m-2})$ where typically, $2 \leq m \leq 5$. 
Further, according to our tests, LNEx needed only up to 650 MB of memory and is able to process, on average, 200 tweets per second.
 
%
%
\section{Experimental Results}

To demonstrate the effectiveness of our context-aware location extractor, we used a set of event-specific hashtags and keywords to collect 4,500 geographically limited, disaster-related tweets from three different targeted streams corresponding to floods in Chennai, Louisiana, and Houston. Below, we categorize and annotate these tweets used for benchmarking each component of LNEx, and for
comparing LNEx with other state-of-the-art tools for the location extraction task.


\subsection{Benchmarking and Annotations}
\label{sec:annotations}


Consistent with the problem of determining whether a location mention is inside the area of interest, our benchmark categorization scheme is not content based as in \cite{matsuda2015annotating,gelernter2013algorithm}.
To better identify and characterize the challenges in extracting location names accurately, our annotation scheme is based on where these locations lie in relation to the area of interest. For example, with respect to Chicago, IL, USA:

\begin{enumerate}[topsep=0pt,noitemsep]
 \item \textbf{$in$\textsc{Loc}}: Locations inside the area of interest, (e.g., Millennium Park or Burlington Ave.) 
 
 \item \textbf{$out$\textsc{Loc}}: Locations outside the area of interest, (e.g., Central Park, 5th Ave, New York.) 
 
 \item \textbf{$amb$\textsc{Loc}}: Ambiguous locations that need context for identification, (e.g., ``our house'')
\end{enumerate}

In contrast to \cite{matsuda2015annotating,gelernter2013algorithm}, our categorization is not based on location types (e.g., buildings, facilities, schools) but on the relative position (i.e., $in$\textsc{Loc} or $out$\textsc{Loc}) and the nature of the location mention (i.e., $in$\textsc{Loc} or $amb$\textsc{Loc}). 
This approach identifies the true scope of challenges in extracting location names. Other schemes that annotate for a limited set of location types, such as ``Geoparse Twitter Benchmark Dataset'', miss obvious location mentions in tweets, such as ``New Zealand'' and ``Christchurch'', making the dataset incompatible for testing the tools mentioned in this paper including LNEx\footnote{Our dataset can be used to test any location extraction tool by ignoring the optional additional expressivity, which makes our dataset more compatible than other available ones.}.
%
%
%
%



\paragraph{\textbf{Tweet Annotations}}

\setlength{\columnsep}{12pt}
\begin{wrapfigure}[3]{r}{0.45\textwidth}
\vspace{-0.5cm}
    \includegraphics[width=0.45\textwidth]{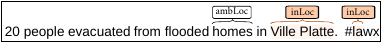}
    \centering
    \caption{Example Annotations using BRAT}
    \label{fig:exampleBrat}
\end{wrapfigure}

Figure \ref{fig:exampleBrat} shows an example of manual annotation from the Louisiana flood tweets using the BRAT tool \cite{bratTool}. It allows us to define search functionalities and additional resources for the annotators to use such as Google Maps.


The annotators annotated three datasets: the 2015 Chennai flood, the 2016 Louisiana flood, and the 2016 Houston flood. In Chennai, they spotted 4,589 location names (75\% $in$\textsc{Loc}, 4\% $out$\textsc{Loc}, and 21\% $amb$\textsc{Loc}); in Louisiana, 2,918 (66\% $in$\textsc{Loc}, 13\% $out$\textsc{Loc}, and 22\% $amb$\textsc{Loc}); and in Houston, 4,177 (66\% $in$\textsc{Loc}, 7\% $out$\textsc{Loc}, and 27\% $amb$\textsc{Loc}). We randomly selected 1k tweets (500 each from Chennai and Louisiana) as a development set and the remaining 3.5k as the test set for evaluation.


\subsection{Evaluation Strategy}

Because BRAT records the start and the end character offsets of the annotated LNs, we evaluate the extraction task by checking the character offsets of the spotted location name in comparison with the annotated data. We used the standard comparison metrics: Precision, Recall, and the balanced F-Score. In the case of overlapping or partial matches, we penalize all tools by adding $\tfrac{1}{2}FP$ (False Positive) and $\tfrac{1}{2}FN$ (False Negative) to the precision and recall equations (e.g., if the tool spots ``The Louisiana'' instead of ``Louisiana'').

We evaluate all tools based on the category of the extracted location in our annotation scheme. For the $in$\textsc{Loc} mentions, we count all hits and misses of a tool and ignore all hits when the category of the extracted location is $out$\textsc{Loc} or $amb$\textsc{Loc}. However, we take a particularly conservative approach and additionally penalize LNEx for extracting location names of $out$\textsc{Loc} and $amb$\textsc{Loc} categories, counting them as false positives (FPs) as our tool is not supposed to extract these.
%
%
\vspace{-0.05cm}

\paragraph{Spell Checking:}

This led to 1\% increase in recall but the F-Score decreased by 2\% on average due to the influence of increased false positives on precision. In the final system, we opted to exclude the spelling corrector component.

\vspace{-0.05cm}


\paragraph{Hashtag Breaking:}

We evaluated the performance of the hashtag breaking component only on the hashtags that contain locations. The accuracies were 97\%, 87\%, and 93\% for Chennai, Louisiana, and Houston respectively, reduced due to examples such as, ``\#lawx'' which was broken into ``law'' and ``x''. 

\vspace{-0.05cm}

\paragraph{Picking a Gazetteer:}

\begin{figure}
    \centering
    \includegraphics[width=1\textwidth]{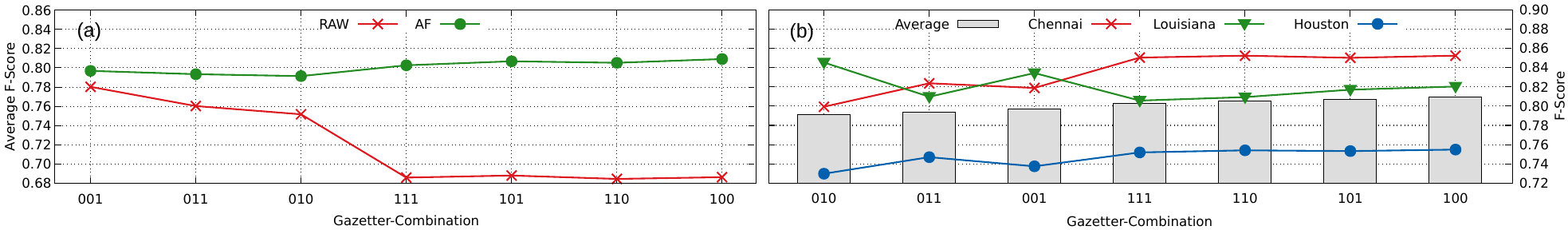}
    \caption[Two numerical solutions]{(a) Raw vs. augmented and filtered ({\scriptsize AF}) gazetteers combinations. (b) Combinations Performance. Each of the seven combinations is a subset of $\{\text{OSM}\;\;\text{Geonames}\;\;\text{DBpedia}\}$.}
    \label{fig:gazzetts}
\end{figure}

The augmentation and filtering of gazetteers improved the F-Scores (See Figure \ref{fig:gazzetts}-a). After this process, combinations of gazetteer sources had similar performance 
where the difference between the worst and the best was around 0.02 F-Score units (see Figure \ref{fig:gazzetts}-b). In the final system, we relied on OSM, which performed the best. Moreover, DBpedia is not focused on geographical information; therefore, it does not contain the metadata useful for the system's future use (e.g., extents and full addresses). Also, OSM has more fine-grained locations and more accurate geo-coordinates than Geonames \cite{gelernter2013automatic}.

\vspace{-0.05cm}

\paragraph{Comparing LNEx with other tools:}

We compared LNEx with the following tools
:

\begin{enumerate}[topsep=0pt,noitemsep]
 \item \textbf{Commercial Grade}: Google NL\footnote{\url{https://cloud.google.com/natural-language/}}, OpenCalais\footnote{\url{http://www.opencalais.com/}}, and Yahoo! BOSS PlaceFinder\footnote{\url{https://developer.yahoo.com/boss/geo/}}. All of these tools have REST APIs and are black box tools that use Machine Learning.

\item \textbf{General Purpose NER}: Stanford NER (SNER) 
and OpenNLP Name Finder. 
SNER learns a linear chain Conditional Random Field (CRF) sequence model \cite{finkel2005incorporating}, while OpenNLP uses the maximum entropy (ME) framework \cite{bender2003maximum}. We trained both tools interchangeably on our annotated datasets in addition to all the data from W-NUT '16\footnote{\url{http://noisy-text.github.io/2016}} while retaining the annotations and unifying the classes we consider as locations (i.e., geo-loc, company, facility) into one type. We used LNEx-OSM gazetteer's features while training SNER. 
Additionally, we used DBpedia Spotlight \cite{mendes2011dbpedia}.

\item \textbf{Twitter NLP}:
OSU Twitter NLP \cite{ritter2011named} 
and TwitIE-Gate \cite{bontcheva2013twitie}. 
Both are pipelined systems of POS-tagging followed by NER. 
TwitIE-GATE also supports normalization, gazetteer lookup, and regular expression-based tagging. For fair comparison, we also augmented them with LNEx OSM gazetteers.

\item \textbf{Twitter Location Extraction}:
Geolocator 3.0 \cite{gelernter2013cross} 
and Geoparsepy \cite{middleton2014real}. 
Geolocator 3.0 uses a tweet-trained CRF classifier and other rule-based models to extract street names, building names, business names, and unnamed locations (i.e., location names containing a category such as ``School'').  

\end{enumerate}

\setlength{\columnsep}{15pt}
\begin{wraptable}[9]{r}{0.49\textwidth}
\vspace{-0.49cm}
\def\arraystretch{1.5}
\tiny
 \centering
 \begin{tabular}{ |l|l| }
    \hline
    Google NLP
    & Location, Organization \\ \hline
    \multirow{2}{*}{OpenCalais
    } & City, Company, Continent, Country, Facility, \\ 
    & Organization, ProvinceOrState, Region, TVStation \\ \hline
    DBpedia Spotlight & Place, Organization \\ \hline
    OSU TwitterNLP & Geo-Location, Company, Facility \\ \hline
    TwitIE-Gate & Location, Organization \\ \hline
    \end{tabular}
    \vspace{7px}
    \caption{Types considered as Locations per tool}
\label{tbl:locTypesPerTool}
\end{wraptable}


All tools have been evaluated using the same metrics and on the same annotated data. In the case of hashtags, we count all hits for all tools and when a tool missed, we penalized only the ones that were  designed to break hashtags (namely, TwitIE-Gate 
and LNEx). 


\setlength{\columnsep}{15pt}
\begin{wrapfigure}[9]{r}{0.49\textwidth}
    \vspace{-1.2cm}
    \includegraphics[width=0.49\textwidth]{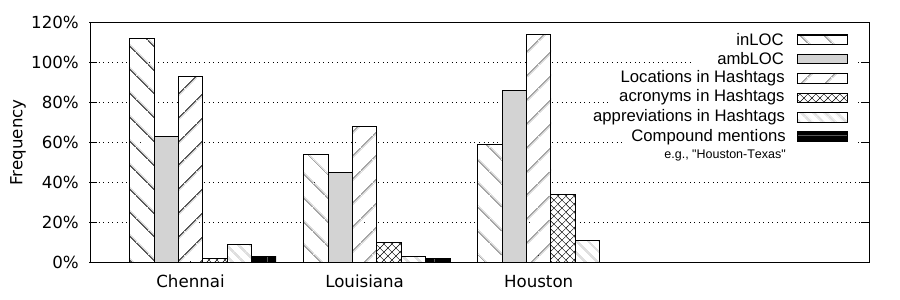}
    \centering
    \caption{Random Sample Evaluation.}
    \label{fig:random_sample_percentages}
\end{wrapfigure}


Additionally, we consider all spotted mentions from PlaceFinder, Geolocator 3.0, Geoparsepy, SNER, and OpenNLP as location names. 

As for the other tools, we consider only the entity types in Table \ref{tbl:locTypesPerTool} as locations.


\setlength{\columnsep}{15pt}
\begin{wraptable}[9]{r}{0.49\textwidth}
\vspace{-2cm}
\def\arraystretch{1.5}
\setlength{\tabcolsep}{0.24em}
\tiny
 \centering
 \begin{tabular}{ l|l|l|l|l|l|l|l|l|l|l }
    \cline{2-11}
    Datasets & \multicolumn{3}{c|}{Chennai} & \multicolumn{3}{c|}{Louisiana} & \multicolumn{3}{c|}{Houston} & \multicolumn{1}{c|}{AVG} \\ \cline{2-11}
    & P & R & F & P & R & F & P & R & F & \multicolumn{1}{c|}{F} \\ \hline
    \multicolumn{1}{|c|}{Google NLP}            & 0.40 & 0.49 & 0.44 & 0.55 & 0.75 & 0.64 & 0.39 & 0.51 & 0.44  & \multicolumn{1}{c|}{0.51} \\ \hline
    \multicolumn{1}{|c|}{OpenCalais}            & 0.43 & 0.10 & 0.17 & 0.81 & 0.77 & 0.78 & 0.62 & 0.35 & 0.45  &  \multicolumn{1}{c|}{0.47} \\ \hline
    \multicolumn{1}{|c|}{DBpedia Spotlight}     & 0.31 & 0.44 & 0.36 & 0.57 & \textbf{0.88} & 0.70 & 0.35 & 0.53 & 0.42  &  \multicolumn{1}{c|}{0.50} \\ \hline
    \multicolumn{1}{|c|}{Yahoo! PLaceFinder}    & 0.67  & 0.39  & 0.49  & \underline{0.83}  & 0.80  & \underline{0.81}  & 0.64  & 0.42  & 0.50  & \multicolumn{1}{c|}{0.61} \\ \hline
    \multicolumn{1}{|c|}{Stanford NER}  & 0.72  & 0.29  & 0.41  & 0.78  & 0.42  & 0.55  & 0.74  & 0.32  & 0.45  & \multicolumn{1}{c|}{0.47} \\ \hline
    \multicolumn{1}{|c|}{OpenNLP}               & 0.55 & 0.15  & 0.24   & 0.62  & 0.19  & 0.29  & 0.60  & 0.23  & 0.34  & \multicolumn{1}{c|}{0.29} \\ \hline
    \multicolumn{1}{|c|}{OSU TwitterNLP}        & 0.74 & 0.40 & 0.52 & \textbf{0.84} & 0.69 & 0.76 & \underline{0.66} & 0.39 & 0.49 & \multicolumn{1}{c|}{0.59} \\ \hline
    \multicolumn{1}{|c|}{TwitIE-Gate}           & 0.51 & 0.36 & 0.43 & 0.66 & \underline{0.84} & 0.74 & 0.35 & 0.39 & 0.37 & \multicolumn{1}{c|}{0.52} \\ \hline
    \multicolumn{1}{|c|}{Geolocator 3.0}        & 0.43 & 0.54 & 0.48 & 0.32 & 0.71 & 0.44 & 0.38 & 0.58 & 0.46 & \multicolumn{1}{c|}{0.46} \\ \hline
    \multicolumn{1}{|c|}{Geoparsepy}        & 0.41 & 0.28 & 0.33 & 0.45 & 0.72 & 0.55 & 0.44 & 0.46 & 0.45 & \multicolumn{1}{c|}{0.45} \\ \hline
    \multicolumn{1}{|c|}{\cellcolor{gray!50} LNEx-RawGaz} & \underline{0.80} & \underline{0.78} & \underline{0.79} & 0.51 & 0.80 & 0.62 & 0.63  & \underline{0.66} & \underline{0.64} & \multicolumn{1}{c|}{\underline{0.69}} \\ \hline
    \multicolumn{1}{|c|}{\cellcolor{black} \textbf{\textcolor{white}{LNEx-AFGaz}}} & \textbf{0.91} & \textbf{0.80} & \textbf{0.85} & \underline{0.83} & 0.81 & \textbf{0.82} & \textbf{0.87} & \textbf{0.67} & \textbf{0.76} & \multicolumn{1}{c|}{\textbf{0.81}} \\ \hline
  \end{tabular}
  \caption{Tools vs. LNEx with a raw (RawGaz) or augmented and filtered gazetteer (AFGaz).}
\label{tbl:unsupervisedToolsScores}
\end{wraptable}

Fig. \ref{fig:random_sample_percentages}) shows that the prevalence of various challenges differ in the three corpora. 
%
%
Nevertheless,  LNEx outperformed all other tools on all datasets in terms of F-Score, and the average F-Score (see Table \ref{tbl:unsupervisedToolsScores}). 
LNEx showed stability on the test and development sets from Louisiana with only a 0.2\% F-Score reduction and around a 2.6\% reduction on the test set from Chennai.

The augmentation and filtering method significantly improved the average F-Score from 0.69 to 0.81. However, limitations of the gazetteer augmentation and filtering methods did contribute to lowering precision. For example, on average, around 5\% of the extracted location names were $out$\textsc{Loc} and $amb$\textsc{Loc}, mistakenly extracted from Chennai, Louisiana, and Houston tweets. Example errors include the augmentation of location names such as ``The $x$ Apartments'' to ``The Apartments'', causing LNEx to extract the phrase ``The Apartments'' as an actual full location name. Fixing such limitations should contribute to around 2\% F-Score improvement on average. 



We trained Stanford NER (SNER) and OpenNLP to emulate their use in other studies mentioned in Section \ref{sec:relatedwork}. Their performances were calculated by interchangeably training them using three datasets at a time and testing on the fourth one (the gazetteer of the area of the test data was also used in training the SNER models). We always used the W-NUT '16 dataset to train the models with more than 10k tweets each time.
%
%

We observed that the ill-formatted text of tweets with ungrammatical text and missing orthographic features impact the F-Score of tools we compared with LNEx. While the performance of each tool differs, we observed that Google heavily relies on orthographic features and expects grammatical texts (although it scored a 0.38 average F-Score). Additionally, TwitIE-GATE was not always successful in extracting location names from hashtags or text even if they are part of the gazetteers that we added to the tool. Finally, OpenCalais extracts only well-known location names of coarser granularity than street and building levels unless a location has an attached location category (e.g., school or street). 

%
%


\setlength{\columnsep}{15pt}
\begin{wraptable}[36]{r}{0.48\textwidth}
\vspace{-0.25cm}
\def\arraystretch{1.5}
\tiny
 \setlength{\tabcolsep}{0.2em}
 \centering
 \begin{tabular}{ |p{0.063\textwidth}|l| }
    \hline
    \multirow{3}{0.5cm}{Original Text}  & sou th kr koil street near Oxford school.west mambalam.. \\ \cline{2-2}
                                        & We r lucky where I am in New Iberia. \#PrayForLouisiana \#lawx \\ \cline{2-2}
                                        & Didn't Houston have a bad flood last year now again  poor htown \\ \hline \hline
    \multirow{3}{1cm}{Manual \\ Annotations \& Types}& (\textbf{$\overbrace{\text{sou th}}^{\text{misspelling}} \underbrace{\text{kr}}_{\text{T6}}$ koil street}) near $\overbrace{\text{(\textbf{Oxford school}).(\textbf{west mambalam})}}^{\text{T6}}$.. \\ \cline{2-2}
                                        & We r lucky where I am in $\overbrace{\text{(\textbf{New Iberia})}}^{\text{T1}}$. $\overbrace{\text{\#PrayFor(\textbf{Louisiana})}}^{\text{T3}}$ $\overbrace{\#(\underbrace{\textbf{\text{la}}}_{\text{T5}})\text{wx}}^{\text{T4}}$ \\ \cline{2-2}
                                        & Didn't $\underbrace{\textbf{\textbf{\text{(Houston)}}}}_{\text{T1}}$ have a bad flood last year now again  poor $\underbrace{\textbf{\textbf{\text{(htown)}}}}_{\text{T5}}$ \\ \hline \hline

    \multirow{3}{0.5cm}{Google NLP}     & sou th kr (\textbf{koil street}) near (\textbf{Oxford}) school.west (\textbf{mambalam}).. \\ \cline{2-2}
                                        & We r lucky where I am in (\textbf{New Iberia}). \#PrayForLouisiana \#lawx \\ \cline{2-2}
                                        & Didn't (\textbf{Houston}) have a bad flood last year now again  poor htown \\ \hline \hline
    \multirow{3}{0.5cm}{OpenCalais}     & sou th kr koil street near (\textbf{Oxford school}).west mambalam.. \\ \cline{2-2}
                                        & We r lucky where I am in New Iberia. \#PrayForLouisiana \#lawx \\ \cline{2-2}
                                        & Didn't (\textbf{Houston}) have a bad flood last year now again  poor htown \\ \hline \hline
    \multirow{3}{0.5cm}{DBpedia Spotlight} & sou th kr koil street near (\textbf{Oxford}) school.west (\textbf{mambalam}).. \\ \cline{2-2}
                                        & We r lucky where I am in (\textbf{New Iberia}). \#PrayForLouisiana \#lawx \\ \cline{2-2}
                                        & Didn't (\textbf{Houston}) have a bad flood last year now again  poor htown \\ \hline \hline
    \multirow{3}{0.5cm}{Yahoo! PlaceFinder}     & sou (\textbf{th}) kr koil street near (\textbf{Oxford}) school.west mambalam.. \\ \cline{2-2}
                                        & We r lucky where I am in (\textbf{New Iberia}). \#PrayForLouisiana \#lawx \\ \cline{2-2}
                                        & Didn't Houston have a bad flood last year now again  poor htown \\ \hline \hline
    \multirow{3}{0.5cm}{Stanford NER}   &  sou th kr koil street near Oxford school.west mambalam..\\ \cline{2-2}
                                            & We r lucky where I am in (\textbf{New Iberia}). \#PrayForLouisiana \#lawx \\ \cline{2-2}
                                            & Didn't Houston have a bad flood last year now again  poor htown \\ \hline \hline
    \multirow{3}{0.5cm}{OpenNLP}        &  sou th kr (\textbf{koil street}) near (\textbf{Oxford}) school.west mambalam.. \\ \cline{2-2}
                                        & We r lucky where I am in (\textbf{New Iberia.}) \#PrayForLouisiana \#lawx \\ \cline{2-2}
                                        & Didn't Houston have a bad flood last year now again  poor htown \\ \hline \hline
    \multirow{3}{1cm}{OSU TwitterNLP}   & sou th kr koil street near (\textbf{Oxford}) school.west mambalam.. \\ \cline{2-2}
                                        & We r lucky where I am in (\textbf{New Iberia}). \#PrayForLouisiana \#lawx \\ \cline{2-2}
                                        & Didn't Houston have a bad flood last year now again  poor htown \\ \hline \hline
    \multirow{3}{1.5cm}{TwitIE-Gate}    & sou th kr koil (\textbf{street}) near (\textbf{Oxford}) school.(\textbf{west mambalam}).. \\ \cline{2-2}
                                        & We r lucky where I am in New Iberia. \#PrayForLouisiana \#lawx \\ \cline{2-2}
                                        & Didn't Houston have a bad flood last year now again  poor htown \\ \hline \hline
    \multirow{3}{0.5cm}{Geolocator 3.0} & (\textbf{sou th}) (\textbf{kr}) (\textbf{koil}) street near (\textbf{Oxford school}).(\textbf{west mambalam}).. \\ \cline{2-2}
                                        & We r lucky where I am in (\textbf{New Iberia}). \#PrayForLouisiana \#lawx \\ \cline{2-2}
                                        & Didn't (\textbf{Houston}) have a bad flood last year now again  poor (\textbf{htown}) \\ \hline \hline
    \multirow{3}{0.5cm}{Geoparsepy}     & sou (\textbf{th}) (\textbf{kr}) koil street near (\textbf{Oxford}) school.west mambalam.. \\ \cline{2-2}
                                        & We r lucky where I am in (\textbf{New (Iberia)}). \#PrayForLouisiana \#lawx \\ \cline{2-2}
                                        & Didn't (\textbf{Houston}) have a bad flood last year now again  poor htown \\ \hline \hline
    \cellcolor{black} & sou th kr koil street near (\textbf{Oxford school}).(\textbf{west mambalam}).. \\ \cline{2-2}
    \cellcolor{black} \textcolor{white}{\textbf{LNEx}} & We r lucky where I am in (\textbf{New Iberia}). \#PrayFor(\textbf{Louisiana}) \#lawx \\ \cline{2-2}
    \cellcolor{black} & Didn't (\textbf{Houston}) have a bad flood last year now again  poor htown \\ \hline
\end{tabular}
\caption{Example tool outputs: bracketed bold text are the identified LNs and braces highlights the types from Fig. \ref{fig:random_sample_percentages}.}
\label{tbl:exampleProblems}
\end{wraptable}


\vspace{-0.25cm}

\paragraph{Illustrative Examples:}

Table \ref{tbl:exampleProblems} shows the comparative handling of three tweets one each from Chennai, Louisiana, and Houston datasets, covering most challenges by all the tools. The location name ``Oxford school'' allowed us to examine if a tool relies on capitalization for delimitation. Only OpenCalais, Geolocator and LNEx were able to extract the name correctly while the rest either partially extracted it or missed it. For example, PlaceFinder extracted ``Oxford'' and geocoded it with the geocodes of Oxford city in England. Although SNER and OpenNLP were trained on the same datasets, OpenNLP extracted Oxford while SNER did not, which suggests that the cue word ``near'' was insufficient evidence for SNER to spot at least Oxford. Correspondingly, since ``New Iberia'' is a correctly capitalized full location name, almost all tools were able to extract it. However, TwitIE-Gate missed it although it is part of the gazetteer we added to the tool, and Geoparsepy extracted Iberia in addition to the full mention, not favoring the longest mention as LNEx. OpenCalais is a black box so we don't know why it failed.

Regarding T3-T5 annotations, LNEx and TwitIE-Gate are designed to break hashtags but TwitIE-Gate was not able to extract any locations from the hashtags in the table. LNEx extracted ``Louisiana'' but was not able to extract ``la'' from ``\#lawx'' due to the statistical method which broke the hashtag into ``law'' and ``x'' since this combination is more probable. Only Geolocator was able to extract the Houston nickname ``htown''. In the future, a dictionary of region-specific acronyms, abbreviations, and nicknames can augment LNEx's region-specific gazetteers.

Google NLP does not handle T6. Adding space between the dot and ``west'' to create ``$\dots$ school. west $\dots$'', results in the  extraction of ``west mambalam'' but omits ``Oxford school''. Google NLP relies on capitalization and so that changing the case of ``s'' to create ``Oxford School'' does help. OpenCalais cannot extract ``west mambalam'' despite fixing all grammatical mistakes, normalizing the orthographic features, and even introducing cue words. The tool only extracts well-known location names of coarser granularity than street and building levels unless they have an attached location category (e.g., school or street). PlaceFinder, on the other hand, tries to find geocodable location names in text. Therefore, the tool extracts ``th'' as the country code of Thailand and ``Oxford'' as the city in England. Hence, geocoding is influencing some of the mistakes of the tool.




\section{Related Work}
\label{sec:relatedwork}


Twitter messages (tweets) lack features exploited by main stream NLP tools. Informality, ill-formed words, irregular syntax and non-standard orthographic features of tweets challenge such tools \cite{kaufmann2010syntactic}. We agree with \cite{baldwin2013noisy} that some issues might be exaggerated. Indeed we found that spelling corrections only contributed to 1\% recall improvement. Nevertheless, text normalization alone is insufficient for NER \cite{derczynski2015analysis}. Specially designed tools such as \cite{ritter2011named,gelernter2013cross} use pipelined systems of POS tagging followed by NER. The latter also perform Regex tagging, normalization, and gazetteer lookup.

Relying on the orthographic features for POS tagging or Regex tagging, previous methods extract locations from the text chunks and phrases of sentences using the following techniques:

\begin{enumerate}[topsep=0pt,noitemsep]
  \item \textbf{Gazetteer search or $n$-gram matching}: \newcite{li2014effective} and \newcite{gelernter2013cross} use a gazetteer matching technique that relies on a segment-based inverted index. \newcite{sultanik2012rapid} use an exhaustive $n$-gram technique. \newcite{middleton2014real} use location-specific gazetteers for matching phrases from tweets. TwitIE-GATE uses a gazetteer lookup component. All of these techniques do not deal with the important issue of the gazetteers' auxiliary content and noise.

  \item \textbf{Handcrafted rules}: \newcite{weissenbacher2015knowledge} and \newcite{malmasi2015location} use pattern and Regex matching which rely on cue words or orthographic features for POS-tagging. TwitIE-GATE adapts rules from ANNIE \cite{cunningham2002gate} for extraction.

  \item \textbf{Supervised Methods}: \textit{Tweet-trained models}: The majority of the methods trained SNER on tweets \cite{gelernter2013cross,yin2014pinpointing} 
  or retrained OpenNLP \cite{lingad2013location}. \textit{News-trained models}: \newcite{malmasi2015location} use tools like SNER and OpenNLP.

  \item \textbf{Semi-supervised methods}: \newcite{ji2016joint} use beam search and structured perceptron for extraction and linking to Foursquare entities. However, they did not address the noise that is prevalent in such sources (e.g., ``my sofa'' or ``our house'') \cite{dalvi2014deduplicating}.
\end{enumerate}



The closest works to ours are TwiNER \cite{li2012twiner} and LEX \cite{downey2007locating}. Both use Microsoft Web $n$-grams (which capture language statistics) for chunking but the former uses DBpedia for entity linking. However, our method exploits a region-specific gazetteer for delimitation and linking. Moreover, LEX worked with web data and relies heavily on capitalization.

Finally, few other methods extract locations from hashtags. \newcite{malmasi2015location} uses a statistical hashtag breaker technique similar to ours. \newcite{ji2016joint} removes only the \# symbol and treats the hashtag as a unigram. \newcite{yin2014pinpointing} uses a greedy maximal matching method for breaking. TwitIE-GATE uses two methods for hashtag breaking: a dynamic programming-based method for finding subsequences and a camel-case-based method for tokenization.


\section{Conclusions and Future Work}



LNEx accurately spots locations in text relying solely on statistical language models synthesized from augmented and filtered region-specific gazetteers. It outperforms state-of-the-art techniques and mainstream location name extractors. By exploiting the knowledge in the gazetteer, we retain the benefits of $n$-gram matching to access location metadata. LNEx does not employ any training and does not depend on syntactic analysis or orthographic conventions. We compensate for limitations in fixed phrase matching with gazetteer augmentation and filtering. Although we do not solve the disambiguation problem here, still the geo/geo ambiguity is reduced by preserving the spatial context through location-specific gazetteers. Furthermore, systematic gazetteer augmentation ties legitimate variants to known locations, minimizing potential ambiguity. 


Certainly, LNEx does not solve all location extraction problems. As the method is driven by the linking procedure, it does not extract location names missing from gazetteers (e.g., ``our house''). It actually presents an effective precision-recall trade-off apparent in the F-Score. In the future, a more sophisticated name model that ignores the generic parts and retains the specific parts when augmenting a location name (e.g., adding ``Sam's'' as a variant of ``Sam's Club'') can be used \cite{dalvi2014deduplicating}.



\section*{Acknowledgments}
This research was partially supported by the NSF award EAR-1520870 ``Hazards SEES: Social and Physical Sensing Enabled Decision Support for Disaster Management and Response''. We would like to also thank Jibril Ikharo for introducing us to the Nameheads work and our other colleagues from Kno.e.sis for helping us in data annotation.

\bibliography{references}
\bibliographystyle{acl}

\end{document}